\documentclass[10pt,letterpaper]{article}

\usepackage{ccn}
\usepackage{pslatex}
\usepackage{apacite}

\usepackage{graphicx}
\usepackage{amsmath}
\usepackage{amssymb}
\usepackage{float}
\usepackage{blkarray}
\usepackage{footnote}
\usepackage{multirow}
\usepackage{natbib}

\usepackage[hyphens]{url}

\usepackage{caption}
\usepackage{subcaption}

\title{Mod-DeepESN: Modular Deep Echo State Network}
 
\author{{\large \bf \shortstack{Zachariah Carmichael\\(zjc2920@rit.edu)} \qquad \shortstack{Humza Syed\\(hxs7174@rit.edu)} \qquad \shortstack{Stuart Burtner\\(stb3444@rit.edu)} \qquad \shortstack{Dhireesha Kudithipudi\\(dxkeec@rit.edu)}} \vspace{1mm} \\
  Neuromorphic AI Lab, Rochester Institute of Technology\\
  1 Lomb Memorial Drive, Rochester, NY 14623 USA
}

\begin{document}
\maketitle
\section{Abstract}
{
\bf
Neuro-inspired recurrent neural network algorithms, such as echo state networks, are computationally lightweight and thereby map well onto untethered devices. The baseline echo state network algorithms are shown to be efficient in solving small-scale spatio-temporal problems. However, they underperform for complex tasks that are characterized by multi-scale structures. In this research, an intrinsic plasticity-infused
modular deep echo state network architecture is proposed to solve complex and multiple timescale temporal tasks.
It outperforms state-of-the-art for time series prediction tasks.}
\begin{quote}
\small
\textbf{Keywords:} 
echo state networks (ESN); intrinsic plasticity; time series prediction; reservoir computing (RC)
\end{quote}
\section{Introduction}
    Echo state networks (ESNs) are recurrent rate-based networks that are efficient in solving spatio-temporal tasks~\citep{jaeger_optimization_2007}. Studies of the associated spiking models have shown that the recurrent layer (\textit{a.k.a.} reservoir layer) corresponds to the granular cells in the cerebellum and the readout layer corresponds to the Purkinjee cells \citep{cereb_2007}. These models are attractive because they offer lightweight, resilient, and conceptually simple networks with rapid training time, as training occurs only at the readout layer. The recurrent nature of these models introduces feedback signals similar to an innate form of fading memory.

	In a recent resurgence to enhance the capabilities of ESNs, few research groups have studied deep ESNs. The premise of adding depth to an ESN is to support hierarchical representations of the temporal input and also capture the multi-scale dynamics of the input features. In prior literature, ESNs have been applied to several temporal tasks, such as speech processing, EEG classification, and anomaly detection \citep{soures2017, jaeger_optimization_2007}. The Deep-ESN architecture proposed by \citeauthor{ma_hier_res_2017}~(\citeyear{ma_hier_res_2017}) consists of stacked reservoir layers and unsupervised encoders to efficiently exploit the temporal kernel property of each reservoir. In \citeauthor{gallicchio_deep_2017}~(\citeyear{gallicchio_deep_2017}), multiple deep ESN architectures based on the shallow Leaky-Integrator ESN of \citeauthor{jaeger_optimization_2007}~(\citeyear{jaeger_optimization_2007}) are analyzed. Of the architectures, the DeepESN network achieved the best results with and without utilizing intrinsic plasticity on memory capacity experiments.

Previous works have shown that increasing the depth of reservoir networks is necessary to capture multi-scale dynamics of time series data as well as extract features with a higher order of complexity \citep{gallicchio_deep_2017}. However, to extract richer features from the data, a wider architecture comprises greater reservoir real estate as each reservoir learns \textit{distinct local features}.
Therefore, we propose a modular deep ESN architecture, known as Mod-DeepESN, which utilizes multiple reservoirs with heterogeneous topology and connectivity to capture and integrate the multi-scale dynamical states of temporal data. The proposed architecture is studied for a mix of benchmarks and consistently performs well across different topologies when compared to the baseline.

\begin{figure}
\centering
  \includegraphics[width=0.81\linewidth]{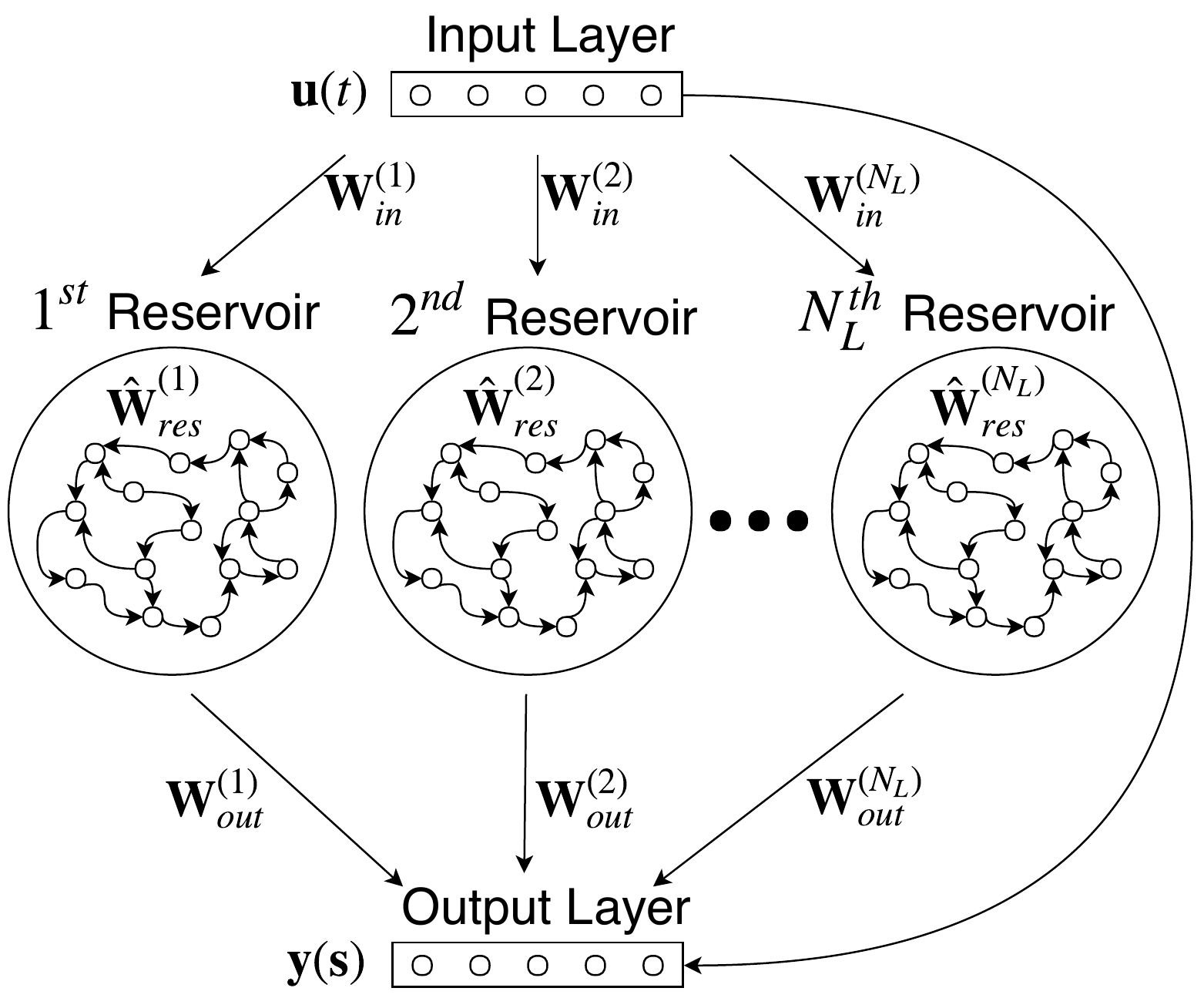}
  \vspace{-0mm}
  \caption{High-level representation of the Mod-DeepESN architecture with the~\textit{Wide} topology.}
  \vspace{-4mm}
  \label{fig:wide}
\end{figure}
\section{Proposed Design}
\label{sec:meth}
\subsection{Architecture}
The proposed Mod-DeepESN architecture consists of modular and deep reservoirs that can be realized in multiple topologies and connectivities. Specifically, four topologies are explored in this study: (i) \textit{Wide} (shown in Figure \ref{fig:wide}) (ii) \textit{Layered} (iii) 2$\times$2 \textit{Criss-Cross} and (iv) \textit{Wide+Layered}.
The reservoirs that are connected to the observed input receive the vector $\mathbf{u}(t)\in \mathbb{R}^{N_U}$ at time $t$, which is projected via the input weight matrix $\mathbf{W}_{in} \in \mathbb{R}^{N_R \mathcal{C}(\mathbf{C}_\mathbf{u},1)\times N_U}$ into each reservoir layer, where $\mathcal{C}(x, j) = \sum_{i}[x_i = j]$ and $[...]$ are Iverson brackets.  $N_U$ is the total number of inputs, $N_R$ is the fixed number of neurons within each reservoir, and lastly $N_L$ is the total number of reservoirs. $\mathbf{C}$ is the binary and triangular connectivity matrix that determines the feedforward connections between reservoirs and the input $\mathbf{u}$. For example, a '1' at position $\mathbf{C}_{\mathbf{u}, 2}$ indicates a feedforward connection from the input $\mathbf{u}$ to the second reservoir. The output of the $l^{\textup{th}}$ reservoir, $\mathbf{x}^{(l)}\in\mathbb{R}^{N_R}$, is computed using \eqref{eq:layer_output}
\vspace{-1.25mm}
\begin{equation}\label{eq:layer_output}
\begin{aligned}
\mathbf{x}^{(l)}(t)= & (1-a^{(l)})\mathbf{x}^{(l)}(t-1)+ \\
 					 & a^{(l)}\tanh\left( \mathbf{W}_{res}^{(l)}\mathbf{i}^{(l)}(t) + \hat{\mathbf{W}}_{res}^{(l)}\mathbf{x}^{(l)}(t-1) \right)
\end{aligned}
\end{equation}
\noindent where $\mathbf{i}^{(l)}(t)$ is computed using \eqref{eq:layer_input}.
\vspace{-1.25mm}
\begin{equation}\label{eq:layer_input}
\mathbf{i}^{(l)}(t) =
\begin{cases}
\mathbf{u}(t) & l=1 \\
\mathbf{x}^{(l-1)}(t) & l>1 \\
\end{cases}
\end{equation}
\noindent Equation \eqref{eq:layer_output} is also referred to as the state transition function in the literature. $\mathbf{W}_{res}^{(l)}\in\mathbb{R}^{N_R\times N_R}$ is a feedforward weight matrix that connects two reservoirs in the \textit{Layered}, 2$\times$2 \textit{Criss-Cross}, and \textit{Wide+Layered} topologies, while $\hat{\mathbf{W}}_{res}^{(l)}\in\mathbb{R}^{N_R\times N_R}$ is a recurrent weight matrix that connects intra-reservoir neurons.
The per-layer leaky parameter $a^{(l)}$ controls the leakage rate in a moving exponential average manner. The non-linear activation function for the recurrent neurons in this work is the hyperbolic tangent. Note that the bias vectors are left out of the formulation for simplicity. The state of the Mod-DeepESN network is defined as the concatenation of the output of each reservoir, \textit{i.e.} $\mathbf{x}(t)=(\mathbf{u}(t), \mathbf{x}^{(1)}(t),...,\mathbf{x}^{(N_L)}(t))\in\mathbb{R}^{N_U + N_L N_R}$. The matrix of all states is denoted by $\chi(\mathbf{s})=(\mathbf{x}(0),...,\mathbf{x}({N_t-1}))\in\mathbb{R}^{(N_U + N_L N_R) \times N_t}$ where $N_t$ is the number of time steps in the time series. Finally, the output of the network for the duration $N_t$ is computed as a linear combination of the Mod-DeepESN state matrix using \eqref{eq:output}.
\vspace{-1.25mm}
\begin{equation}\label{eq:output}
\mathbf{y}(\mathbf{s})=\mathbf{W}_{out}\chi(\mathbf{s})
\end{equation}
\noindent The matrix $\mathbf{W}_{out}\in \mathbb{R}^{N_Y\times (N_U + N_L N_R)}$ contains the feedforward weights between reservoir neurons and the $N_Y$ output neurons.
Ridge regression using the Moore-Penrose pseudo inverse is used to solve for optimal $\mathbf{W}_{out}$, shown in (\ref{eq:mppi}).
\vspace{-1.25mm}
\begin{equation}\label{eq:mppi}
\mathbf{W}_{out} = \mathbf{y}(\mathbf{s}) \cdot \chi(\mathbf{s})^\intercal \cdot \left (\chi(\mathbf{s})\cdot\chi(\mathbf{s})^\intercal + \beta \mathbb{I} \right ) ^{-1}
\end{equation}
\noindent The formulation includes a regularization term $\beta$, and $\mathbb{I}$ is the identity matrix. This computation is performed only during the training phase of the model.
\subsection{Mod-DeepESN Topologies}\label{subsec:top}
For all the proposed topologies, (i) \textit{Wide} (ii) \textit{Layered} (iii) 2$\times$2 \textit{Criss-Cross} and (iv) \textit{Wide+Layered}, the input layer is directly connected to the output layer; this connection captures sensitive spatial features and improves the network performance experimentally. Additionally, each reservoir is connected to the output layer.

Figure \ref{fig:wide} illustrates the \textit{Wide} topology. Distinct connections are created between the input layer and each reservoir, \textit{i.e.} the input weights are not shared between reservoirs. As a result, an ensemble of shallow ESNs emerges, each of which captures varying dynamics of the input data locally.

\begin{figure}
\centering
  \includegraphics[width=.85\linewidth]{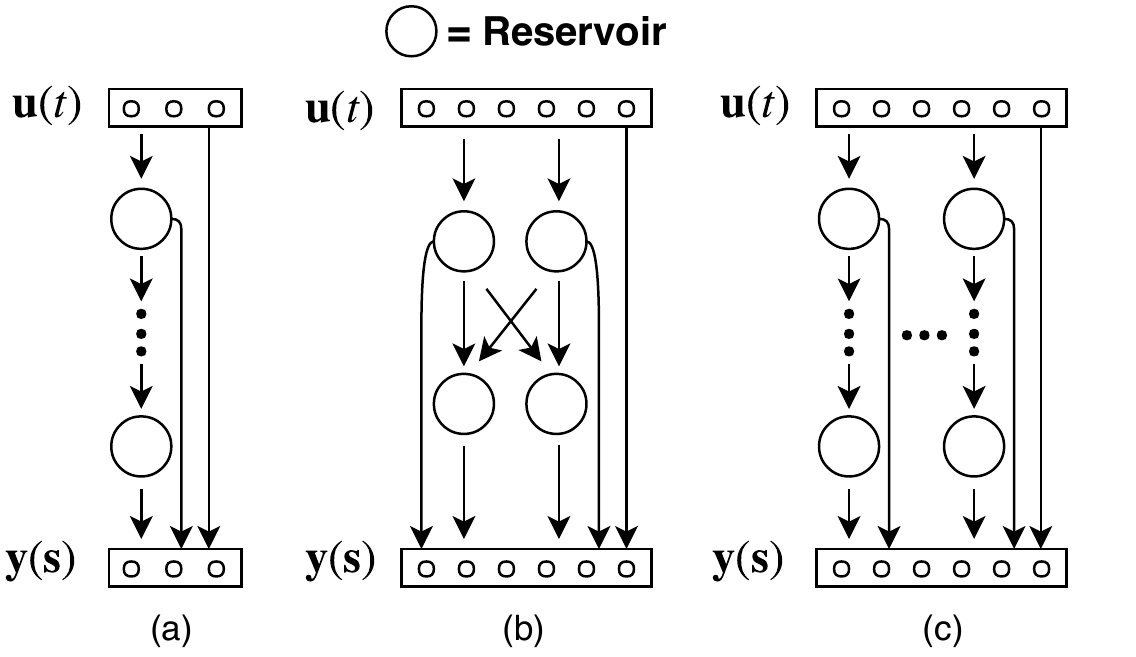}
  \vspace{-0mm}
  \caption{(a) \textit{Layered} topology, (b) 2$\times$2 \textit{Criss-Cross} topology, and (c) \textit{Wide+Layered} topology of Mod-DeepESN.}
  \vspace{-4mm}
  \label{fig:deep_cross}
\end{figure}

The \textit{Layered} topology, shown in Figure \ref{fig:deep_cross}(a), presents the input to the first reservoir with each successive reservoir receiving feedforward input from its predecessor.
By stacking reservoirs in this fashion, stateful dynamics of the system are integrated with features that increase in complexity with depth.

Figure \ref{fig:deep_cross}(b) depicts the 2$\times$2 \textit{Criss-Cross} topology, which comprises a total of 4, or $N_L = N ^ 2$ ($N = 2$), reservoirs. This model exhibits dense feedforward connectivity between all reservoirs, but maintains reservoir-local recurrent connections. The system states are hierarchically integrated at multiple depths from disparate inputs.

The \textit{Wide+Layered} topology, presented in Figure \ref{fig:deep_cross}(c), deviates from the 2$\times$2 \textit{Criss-Cross} topology through greater sparsity in its feedforward connectivity; dynamical states of the input are integrated in discrete pathways that resemble the \textit{Layered} topology.
\subsection{Weight Initialization}\label{subsec:weights}
As weights in ESN reservoirs are fixed, recurrent weights must be initialized to maintain stable reservoir states, especially in deep ESNs where inputs to a network propagate through multiple reservoirs. As mathematically supported in the ESN literature,
deep ESNs need to satisfy the \textit{Echo State Property} for such stability. Accordingly, each $\hat{\mathbf{W}}_{res}^{(l)}$ is initialized using a uniform distribution and scaled such that \eqref{eq:echo_state_property} is satisfied.
\vspace{-1.25mm}
\begin{equation}\label{eq:echo_state_property}
\max_{1\le l \le N_L} {\rho \left( (1-a^{(l)})\mathbf{\mathbb{I}}+a^{(l)}\hat{\mathbf{W}}_{res}^{(l)} \right) < 1}
\end{equation}
\noindent The spectral radius of the matrix, \textit{i.e.} the magnitude of the largest eigenvalue, is computed by the function $\rho(\cdot)$. We substitute a hyperparameter $\hat{\rho}$ for the value 1 to fine-tune model reservoirs for each experiment.

The remaining weight matrices are initialized from a uniform distribution such that the Euclidean distance between weights is equivalent to either $\sigma_{in}$ or $\sigma_{l}$, \textit{i.e.} $||\mathbf{W}_{in}||_2=\sigma_{in}$ and $||\mathbf{W}_{res}^{(l)}||_2=\sigma_{res}$. $\sigma_{in}$ and $\sigma_{l}$ are tunable hyperparameters. 

Rather than normalizing weights using the $L_2$ norm or spectral radius, Xavier initialization \citep{glorot2010understanding} can be used to determine the distribution of weights. In this method, any set of weights $\mathbf{W}$ can be drawn from a normal distribution as shown by (\ref{eq:xav_std}) and (\ref{eq:xav_normal}).
\vspace{-1.25mm}
\begin{gather}
  \sigma_{xav} = \sqrt{{2} / {(n_{in} + n_{out})}}    \label{eq:xav_std} \\
  \mathbf{W} \sim \mathcal{N}(\mu=0, \sigma_{xav}^2)    \label{eq:xav_normal}
\end{gather}
\noindent The distribution is parameterized with a mean of $\mu=0$ and a standard deviation $\sigma_{xav}$, where $n_{in}$ is the number of inputs and $n_{out}$ is the number of outputs that the weight matrix is connecting. Additionally, this initialization can be made modular to isolate which sets of weights utilize a Xavier initialization and which use either $L_2$ or spectral radius normalization.

All modes of initialization additionally use a sparsity hyperparameter that determines the probability that each weight is nullified. Specifically, $s_{in}$ determines sparsity for $\mathbf{W}_{in}$, $\hat{s}_{l}$ for each $\hat{\mathbf{W}}_{res}^{(l)}$, and $s_{l}$ for each $\mathbf{W}^{(l)}_{res}$.
\subsection{Intrinsic Plasticity (IP)}
In this section,  we study how a neurally-inspired IP mechanism can enhance the performance of the Mod-DeepESN.
Originally proposed by \citeauthor{schrauwen_improving_2008}~(\citeyear{schrauwen_improving_2008}), the IP rule introduces a gain and bias term to the $\tanh(\cdot)$ function: $\tanh(x)\rightarrow\tanh(gx+b)$ where $g$ is the gain and $b$ is the bias. The update rules are given by \eqref{eq:ip_b} and \eqref{eq:ip_g} where
\vspace{-1.2mm}
\begin{gather}
  \Delta b = -\eta \left( \left(-\frac{\mu}{\sigma^2}\right) + \left(\frac{\tilde{x}}{\sigma^2}\right) \left( 2 \sigma^2 + 1 - \tilde{x}^2 + \mu \tilde{x} \right) \right)    \label{eq:ip_b} \\
  \Delta g = {\eta}/{g} + \Delta b x    \label{eq:ip_g}
\end{gather}
\noindent $x$ is the net weighted sum of inputs to a neuron, $\tilde{x}$ is the $\tanh$ activation of $x$, \textit{i.e.} $\tilde{x}=\tanh(gx+b)$, and $\eta$ is the learning rate. The hyperparameters $\sigma$ and $\mu$ are the standard deviation and mean, respectively, of the target Gaussian distribution. In a pre-training phase, the learned parameters are initialized as $b=0$ and $g=1$, and are updated iteratively according to \eqref{eq:ip_b} and \eqref{eq:ip_g}. The application of such results in the minimization of the Kullback-Leibler (KL) divergence between the empirical output distribution and the target Gaussian distribution \citep{schrauwen_improving_2008}.
\subsection{Genetic Algorithm}
To fine-tune network hyperparameters, a genetic algorithm is employed \citep{back2000evolutionary}. The evolution is executed with a population size of 50 individuals for 50 generations, and evaluated in a tournament comprising 3 randomly selected individuals. Individuals mate with a crossover probability of 50\% and mutate with a probability of 10\% to form each successive generation. This evolution is used to tune all model hyperparameters as well as the Mod-DeepESN topology type.
\section{Experiments \& Results}
\label{sec:experiment}
Two datasets, the chaotic Mackey-Glass time series generated using the fourth-order Runge-Kutta method (RK4) and a daily minimum temperature series \citep{temp_data}, are used to evaluate Mod-DeepESN as they present variable nonlinearity at multiple time scales. 
    To analyze the effectiveness of the Mod-DeepESN model, root mean squared error (RMSE), normalized RMSE (NRMSE), and mean absolute percentage error (MAPE) are computed as shown in \eqref{eq:rmse}, \eqref{eq:nrmse}, and \eqref{eq:mape}, respectively.
\vspace{-1.25mm}
\begin{gather}
  \textup{RMSE} = \sqrt{\frac{1}{N_t} \sum_{t=1}^{N_t} [\mathbf{u}(t) - \hat{\mathbf{u}}(t)]^2}     \label{eq:rmse} \\
  \textup{NRMSE} = \sqrt{\sum_{t=1}^{N_t} [\mathbf{u}(t) - \hat{\mathbf{u}}(t)]^2 / \left ( \sum_{t=1}^{N_t} [\mathbf{u}(t) - \bar{\mathbf{u}}]^2 \right ) }     \label{eq:nrmse}\\
  \textup{MAPE} = \frac{1}{N_t} \sum_{t=1}^{N_t} \frac{|\mathbf{u}(t) - \hat{\mathbf{u}}(t)|}{\mathbf{u}(t)} \times 100\%     \label{eq:mape}
\end{gather}
\noindent $\hat{\mathbf{u}}(t)$ is the predicted time series value at time step $t$, and $\bar{\mathbf{u}}$ is the average value of the time series over the $N_t$ time steps. Each reported result is averaged over 10 runs using the best hyperparameters found from the genetic algorithm search.

\begin{figure}
  \begin{subfigure}[c]{0.9\linewidth}
    \includegraphics[width=\linewidth]{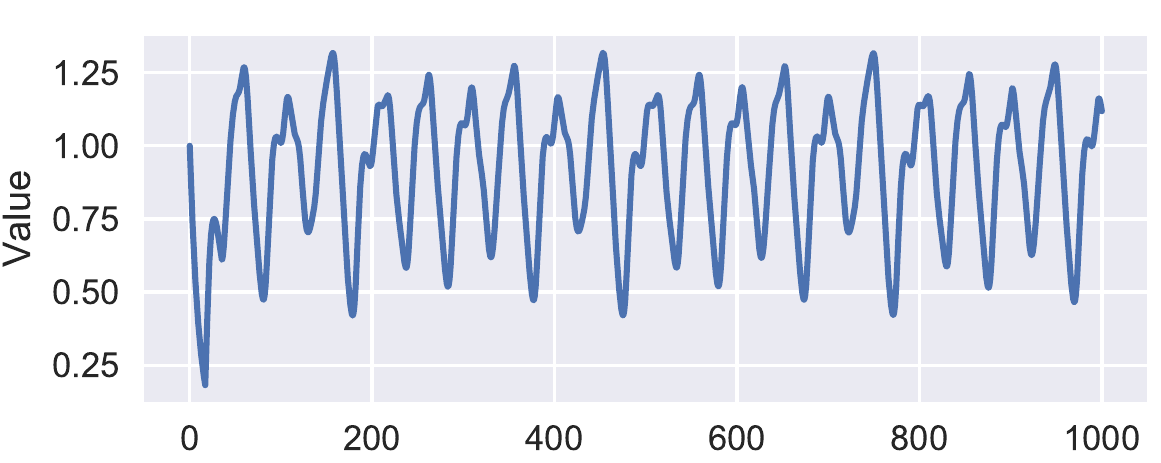}
  \end{subfigure}\hfill
  \begin{subfigure}[c]{0.1\linewidth}
    \caption{}
  \end{subfigure}
  \begin{subfigure}[c]{0.9\linewidth}
    \includegraphics[width=\linewidth]{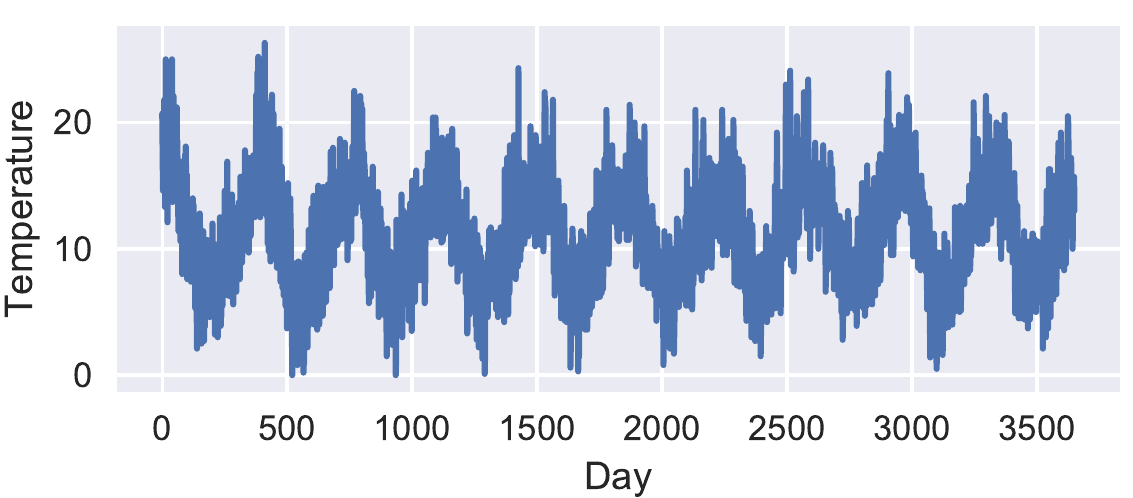}
  \end{subfigure}\hfill
  \begin{subfigure}[c]{0.1\linewidth}
    \caption{}
  \end{subfigure}
  \caption{(a) The Mackey-Glass time series and (b) the Melbourne daily minimum temperature time series.}
  \label{fig:mackey}
  \vspace{-4mm}
\end{figure}

	The Mackey-Glass dataset, shown in Figure \ref{fig:mackey}(a), is split into 8,000 training samples and 2,000 testing samples to forecast 84 time steps in advance. To reduce the influence of initial reservoir states, the first 100 predictions from the network are discarded in a washout period. 
    
    The daily minimum temperature series dataset, shown in Figure \ref{fig:mackey}(b), comprises data from Melbourne, Australia (1981-1990), and is split into 2,920 training samples and 730 testing samples. A washout period of 30 steps is realized during model evaluation for the single-step ahead prediction.
\begin{table*}
  \centering
  \caption{Mackey-Glass Time Series 84-Step Ahead Prediction Results.}\vspace{-2mm}
  {\small
  \label{tab:mg_results}
  \begin{tabular}{|c|c|*{11}{|c}|*{3}{|c}|}
    \hline
    \multicolumn{2}{|c||}{Method} & $N_L$ & $N_R$ & $\beta$ & $\alpha$ & $\hat{\rho}$ & $\sigma_{in}$ & $\sigma_{l}$ & $s_{in}$ & $\hat{s}_{l}$ & $s_{l}$ & IP & RMSE$\times$e-3 & NRMSE$\times$e-3 & MAPE$\times$e-3\\  
    \hline\hline
    \multirow{5}{*}{\hspace{-1.25mm}\rotatebox[origin=c]{90}{Baseline}\hspace{-1.25mm}} &
    ESN        \footnote[1] & 1 & - & - & - & - & - & - & - & - & - & - & 43.7 & 201 & 7.03  \\ 
    &$\phi$-ESN \footnote[2] & 2 & - & - & - & - & - & - & - & - & - & - & 8.60 & 39.6 & 1.00 \\
    &R$^2$SP    \footnote[3]     & 2 & - & - & - & - & - & - & - & - & - & - & 27.2 & 125 & 1.00 \\
    &MESM       \footnote[4] & 7 & - & - & - & - & - & - & - & - & - & - & 12.7 & 58.6 & 1.91 \\
    &Deep-ESN   \footnote[5] & 3 & - & - & - & - & - & - & - & - & - & - & \textbf{1.12} & \textbf{5.17} & \textbf{.151} \\
    \hline
    \multirow{5}{*}{\hspace{-1.25mm}\rotatebox[origin=c]{90}{This Work}\hspace{-1.25mm}} &
    \textit{Wide}          & 3     & \multirow{5}{*}{256} & \multirow{5}{*}{2e-8} & \multirow{5}{*}{.6} & \multirow{5}{*}{$X$} & \multirow{5}{*}{.1} & \multirow{5}{*}{$X$} & \multirow{5}{*}{.1} & \multirow{5}{*}{.1} & \multirow{5}{*}{.7} & N & {20.2} & {48.7} & {8.11} \\
    &\textit{Layered}      & 3     &  &  &  &  &  &  &  &  &  & N & 57.8  & 96.2  & 19.9  \\
    &\textit{Criss-Cross}  & 4 (2) &  &  &  &  &  &  &  &  &  & N & 56.1  & 54.4  & 8.94  \\
    &\textit{Wide+Layered} & 6 (3) &  &  &  &  &  &  &  &  &  & N & 41.1  & 55.4  & 11.2  \\
    &\textit{Wide}         & 3     &  &  &  &  &  &  &  &  &  & Y & {7.22} & {27.5} & {5.55} \\
    \hline
  \end{tabular}\\[-2mm]}
\end{table*}
\begin{table*}
  \centering
  \caption{Daily Minimum Temperature Series 1-Step Ahead Prediction Results.}\vspace{-2mm}
  {\small
  \label{tab:temp_results}
  \begin{tabular}{|c|c|*{11}{|c}|*{3}{|c}|}
    \hline
    \multicolumn{2}{|c||}{Method} & $N_L$ & $N_R$ & $\beta$ & $\alpha$ & $\hat{\rho}$ & $\sigma_{in}$ & $\sigma_{l}$ & $s_{in}$ & $\hat{s}_{l}$ & $s_{l}$ & IP & RMSE$\times$e-3 & NRMSE$\times$e-3 & MAPE$\times$e-3\\
    \hline\hline
    \multirow{5}{*}{\hspace{-1.25mm}\rotatebox[origin=c]{90}{Baseline}\hspace{-1.25mm}} &
    ESN 	   \footnote[1] & 1 & - & - & - & - & - & - & - & - & - & - & 501  & 139  & 39.5  \\ 
    & $\phi$-ESN \footnote[2] & 2 & - & - & - & - & - & - & - & - & - & - & 493  & 141  & 39.6  \\
    & R$^2$SP    \footnote[3] & 2 & - & - & - & - & - & - & - & - & - & - & 495  & 137  & 39.3  \\
    & MESM	   \footnote[4] & 7 & - & - & - & - & - & - & - & - & - & - & 478  & 136  & 37.7  \\
    & Deep-ESN   \footnote[5] & 2 & - & - & - & - & - & - & - & - & - & - & 473  & 135  & \textbf{37.0} \\
    \hline
    \multirow{5}{*}{\hspace{-1.25mm}\rotatebox[origin=c]{90}{This Work}\hspace{-1.25mm}} &
    \textit{Wide} & 2 & \multirow{5}{*}{1024} & \multirow{5}{*}{7e-4} & \multirow{5}{*}{1} & \multirow{5}{*}{$X$} & \multirow{5}{*}{.4} & \multirow{5}{*}{$X$} & \multirow{5}{*}{.6} & \multirow{5}{*}{.3} & \multirow{5}{*}{.6} & N & 473 & 135  & 38.6  \\
    & \textit{Layered} & 2 &  &  &  &  &  &  &  &  &  & N & {470} & {134} & {38.2} \\
    & \textit{Criss-Cross} & 4 (2) &  &  &  &  &  &  &  &  &  & N & 472  & 135  & 38.7  \\
    & \textit{Wide+Layered} & 4 (2) &  &  &  &  &  &  &  &  &  & N & 471  & 135  & 38.6  \\
    & \textit{Wide+Layered} & 4 (2) &  &  &  &  &  &  &  &  &  & Y & \textbf{459} & \textbf{132} & {37.1} \\
    \hline
  \end{tabular}\\[-2mm]}
\end{table*}

In each experiment, $a^{(l)}$ is set to the same value to reduce the hyperparameter search space. We denote this value as $\alpha$ in Tables \ref{tab:mg_results} and \ref{tab:temp_results}. Additionally, a value in the $N_L$ column with parenthesis is used to indicate the reservoir width of the 2$\times$2 \textit{Criss-Cross} and \textit{Wide+Layered} topologies. If Xavier initialization is used in place of initialization by spectral radius or $L_2$ normalization, $X$ is reported. Hyperparameters are chosen independently of the topology, as the experimental performance consistently surpasses that of the best hyperparameters found per topology. Each topology is evaluated with and without IP to determine its efficacy, but only the best result using IP is reported. Baseline ESN results for both datasets are retrieved from \citeauthor{ma_hier_res_2017}~(\citeyear{ma_hier_res_2017}).
The Mod-DeepESN implementation makes use of the scikit-learn \citep{scikit-learn}, SciPy \citep{SciPy}, NumPy \citep{oliphant2006guide}, and DEAP \citep{DEAP_JMLR2012} software libraries.

	The Mackey-Glass experimental results demonstrate that model fitness depends on the Mod-DeepESN topology. An IP pre-training phase greatly reduces the network error and performs best while evaluated with the \textit{Wide} topology. The proposed architecture outperforms all baseline models with the exception of the Deep-ESN \citep{ma_hier_res_2017}.

    In the daily minimum temperature time series experiment, the \textit{Wide+Layered} Mod-DeepESN outperforms every baseline model, and further reduces the error with the inclusion of an IP pre-training phase. Varying the topology within this experiment is less impactful than within the Mackey-Glass experiment, but deeper models consistently yield better performance.
    
    It is important to note that the best Mod-DeepESN models use Xavier initialization for multiple weight matrices. While this does not necessarily suggest that spectral radius or $L_2$ normalization are inferior initialization methods, it does indicate that a desirable initialization can be achieved by using a non-parametrized Gaussian distribution that reduces the hyperparameter search space. Furthermore, the best-performing Mod-DeepESN models, \textit{Wide} and \textit{Wide+Layered}, incorporate distinct, non-coinciding channels from input layer to output layer, and consistently outperform the other topologies.
\begin{savenotes}
\phantom{
  \footnote[1]{\small Leaky-integrator ESN \citep{jaeger_optimization_2007}.}\\
  \footnote[2]{\small $\phi$-ESN \citep{phi_2011}.}\\
  \footnote[3]{\small R$^2$SP \citep{r2sp_2013}.}\\
  \footnote[4]{\small Multi-layered echo state machine (MESM) \citep{mesm_2017}.}\\
  \footnote[5]{\small Deep-ESN \citep{ma_hier_res_2017}.}\\
}
\end{savenotes}
\section{Conclusions}
In this work, we have shown the efficacy of a deep and modular echo state network architecture with varying topology. Prediction error on multiple datasets is reduced by isolating the integration of dynamical input states to disparate computational pathways, and by incorporating IP in a pre-training phase. Xavier initialization of weights decreases the complexity of hyperparameter tuning and performs well consistently across all proposed topologies. By combining these mechanisms with a genetic algorithm, the Mod-DeepESN outperforms several baseline models on non-trivial time series forecasting tasks.

\section{Acknowledgments}
The authors would like to thank the members of the RIT Neuromorphic AI Lab for their valuable feedback on this work.

\Urlmuskip=0mu plus 1mu\relax
\bibliographystyle{apacite}
\setlength{\bibleftmargin}{.125in}
\setlength{\bibindent}{-\bibleftmargin}
\renewcommand*{\bibfont}{\scriptsize}
\bibliography{bibby}

\end{document}